\documentclass[11pt]{article}
\usepackage{coling2018}
\usepackage{times}
\usepackage{url}
\usepackage{multirow}
\usepackage{latexsym}
\usepackage{graphicx}
\usepackage{diagbox}
\usepackage{enumitem} 
\usepackage{color}
\usepackage{subfigure}
\usepackage{footnote}
\makesavenoteenv{tabular}
\makesavenoteenv{table}

\title{Design Challenges and Misconceptions in Neural Sequence Labeling}

\author{Jie Yang, Shuailong Liang, Yue Zhang \\ 
  Singapore University of Technology and Design \\
  {\tt \{jie\_yang, shuailong\_liang\}@mymail.sutd.edu.sg} \\
  {\tt yue\_zhang@sutd.edu.sg} \\}

\date{}

\begin{document}
\maketitle
\begin{abstract}
  We investigate the design challenges of constructing effective and efficient neural sequence labeling systems, by reproducing twelve neural sequence labeling models, which include most of the state-of-the-art structures, and conduct a systematic model comparison on three benchmarks (i.e. NER, Chunking, and POS tagging). Misconceptions and inconsistent conclusions in existing literature are examined and clarified under statistical experiments. In the comparison and analysis process, we reach several practical conclusions which can be useful to practitioners.
\end{abstract}

\section{Introduction} \label{intro}
\blfootnote{This work is licenced under a Creative Commons Attribution 4.0 International Licence. Licence details: \url{http://creativecommons.org/licenses/by/4.0/}}
Sequence labeling models have been used for fundamental NLP tasks such as POS tagging, chunking and named entity recognition (NER). Traditional work uses statistical approaches such as Hidden Markov Models (HMM) and Conditional Random Fields (CRF) \cite{ratinov2009design,passos2014lexicon,luo2015joint} with handcrafted features and task-specific resources. With advances in deep learning, neural models have given state-of-the-art results on many sequence labeling tasks \cite{ling2015finding,lample2016neural,ma2016end}. Words and characters are encoded in distributed representations \cite{mikolov2013distributed}  and sentence-level features are learned automatically during end-to-end training. Many existing state-of-the-art neural sequence labeling models utilize word-level Long Short-Term Memory (LSTM) structures to represent global sequence information and a CRF layer to capture dependencies between neighboring labels \cite{huang2015bidirectional,lample2016neural,ma2016end,peters2017semi}. As an alternative, Convolution Neural Network (CNN) \cite{lecun1989backpropagation} has also been used for its ability of parallel computing, leading to an efficient training and decoding process.

Despite them being dominant in the research literature, reproducing published results for neural models can be challenging, even if the codes are available open source. For example, \newcite{reimers2017reporting} conduct a large number of experiments using the code of \newcite{ma2016end}, but cannot obtain comparable results as reported in the paper. \newcite{liu2017empower} report lower average F-scores on NER when reproducing the structure of \newcite{lample2016neural}, and on POS tagging when reproducing \newcite{ma2016end}. Most literature compares results with others by citing the scores directly \cite{huang2015bidirectional,lample2016neural} without re-implementing them under the same setting, resulting in less persuasiveness on the advantage of their models. In addition, conclusions from different reports can be contradictory. For example, most work observes that stochastic gradient descent (SGD) gives best performance on NER task \cite{chiu2015named,lample2016neural,ma2016end}, while \newcite{reimers2017reporting} report that SGD is the worst optimizer on the same datasets. 

The comparison between different deep neural models is challenging due to sensitivity on experimental settings. We list six inconsistent configurations in literature, which lead to difficulties for fair comparison.

\begin{table*}[!tp]
\begin{center}
\resizebox{\columnwidth}{!}{%
\begin{tabular}{|l|l|l|l|l|}
\hline   
\textbf{Models}&\textbf{Word LSTM+CRF}& \textbf{Word LSTM} &\textbf{Word CNN+CRF} &\textbf{Word CNN}  \\ 
\hline
\multirow{3}*{\textbf{No Char}}
&\newcite{huang2015bidirectional}* &\newcite{ma2016end}  &\newcite{collobert2011natural}* &\newcite{strubell2017fast}* \\
 &\newcite{lample2016neural} &\newcite{strubell2017fast}*  &\newcite{dos2015boosting} &\\
&\newcite{strubell2017fast}* &  &\newcite{strubell2017fast}* &  \\
\hline
\multirow{3}*{\textbf{Char LSTM}} &\newcite{lample2016neural} &\newcite{lample2016neural}& No existing work&No existing work  \\
&\newcite{rei2017semi} &  & &  \\
&\newcite{liu2017empower} &  & &  \\
\hline
\multirow{3}*{\textbf{Char CNN}} &\newcite{ma2016end} &\newcite{ma2016end} &\newcite{dos2015boosting} & \newcite{santos2014learning} \\
&\newcite{chiu2015named}* &   & &   \\
&\newcite{peters2017semi} &   & &   \\
\hline
\end{tabular}
}
\end{center}
\caption{Neural sequence labeling models in literatures. * represents using handcrafted neural features.}
\label{tab:models}
\end{table*}

\noindent \textbullet $\;$ \textbf{Dataset}. Most work reports sequence labeling results on both CoNLL 2003 English NER \cite{tjong2003introduction} and PTB POS \cite{marcus1993building} datasets \cite{collobert2011natural,huang2015bidirectional,ma2016end}. \newcite{ling2015finding} give results only on POS dataset, while some papers \cite{chiu2015named,lample2016neural,strubell2017fast} report results on the NER dataset only. \newcite{dos2015boosting} conducts experiments on NER for Portuguese and Spanish. 

Most work uses the development set to select hyperparameters \cite{lample2016neural,ma2016end}, while others add development set into training set \cite{chiu2015named,peters2017semi}. \newcite{reimers2017reporting} use a smaller dataset (13862 vs 14987 sentences). Different from \newcite{ma2016end} and \newcite{liu2017empower}, \newcite{huang2015bidirectional} choose a different data split on the POS dataset. \newcite{liu2017empower} and \newcite{hashimoto2017joint} use different development sets for chunking.

\noindent \textbullet $\;$ \textbf{Preprocessing}. A typical data preprocessing step is to normize digit characters \cite{chiu2015named,lample2016neural,yang2016multi,strubell2017fast}. \newcite{reimers2017reporting} use fine-grained representations for less frequent words. \newcite{ma2016end} do not use preprocessing.


\noindent \textbullet $\;$ \textbf{Features}. \newcite{strubell2017fast} and \newcite{chiu2015named} apply word spelling features and \newcite{huang2015bidirectional} further integrate context features.  \newcite{collobert2011natural} and \newcite{huang2015bidirectional} use neural features to represent external gazetteer information. Besides, \newcite{lample2016neural} and \newcite{ma2016end} use end-to-end structure without handcrafted features.

\noindent \textbullet $\;$ \textbf{Hyperparameters} including learning rate, dropout rate \cite{srivastava2014dropout}, number of layers, hidden size etc. can strongly affect the model performance. \newcite{chiu2015named} search for the hyperparameters for each task and show that the system performance is sensitive to the choice of hyperparameters. However, existing models use different parameter settings, which affects the fair comparison.

\noindent \textbullet $\;$ \textbf{Evaluation}. Some literature reports results using mean and standard deviation under different random seeds \cite{chiu2015named,peters2017semi,liu2017empower}. Others report the best result among different trials \cite{ma2016end}, which cannot be compared directly.

\noindent \textbullet $\;$ \textbf{Hardware environment} can also affect system accuracy. \newcite{liu2017empower} observes that the system gives better accuracy on NER task when trained using GPU as compared to using CPU. Besides, the running speeds are highly affected by the hardware environment.

To address the above concerns, we systematically analyze neural sequence labeling models on three benchmarks: CoNLL 2003 NER \cite{tjong2003introduction}, CoNLL 2000 chunking \cite{tjong2000introduction} and PTB POS tagging \cite{marcus1993building}. Table \ref{tab:models} shows a summary of the models we investigate, which can be categorized under three settings: (i) character sequence representations ; (ii) word sequence representations; (iii) inference layer. Although various combinations of these three settings have been proposed in the literature, others have not been examined. We compare all models in Table \ref{tab:models}, which includes most state-of-the-art methods. To make fair comparisons, we build a unified  framework\footnote{Our code has been released at \url{https://github.com/jiesutd/NCRFpp}.} to reproduce the twelve neural sequence labeling architectures in Table \ref{tab:models}. Experiments show that our framework gives comparable or better results on reproducing existing works, showing the practicability and reliability of our analysis for practitioners. The detailed comparison and analysis show that (i) Character information provides a significant improvement on accuracy; (ii) Word-based LSTM models outperforms CNN models in most cases; (iii) CRF can improve model accuracy on NER and chunking but does not on POS tagging. Our framework is based on PyTorch with batched implementation, which is highly efficient, facilitating quick configurations for new tasks.

\begin{figure} 
  \centering 
    \includegraphics[width=5in]{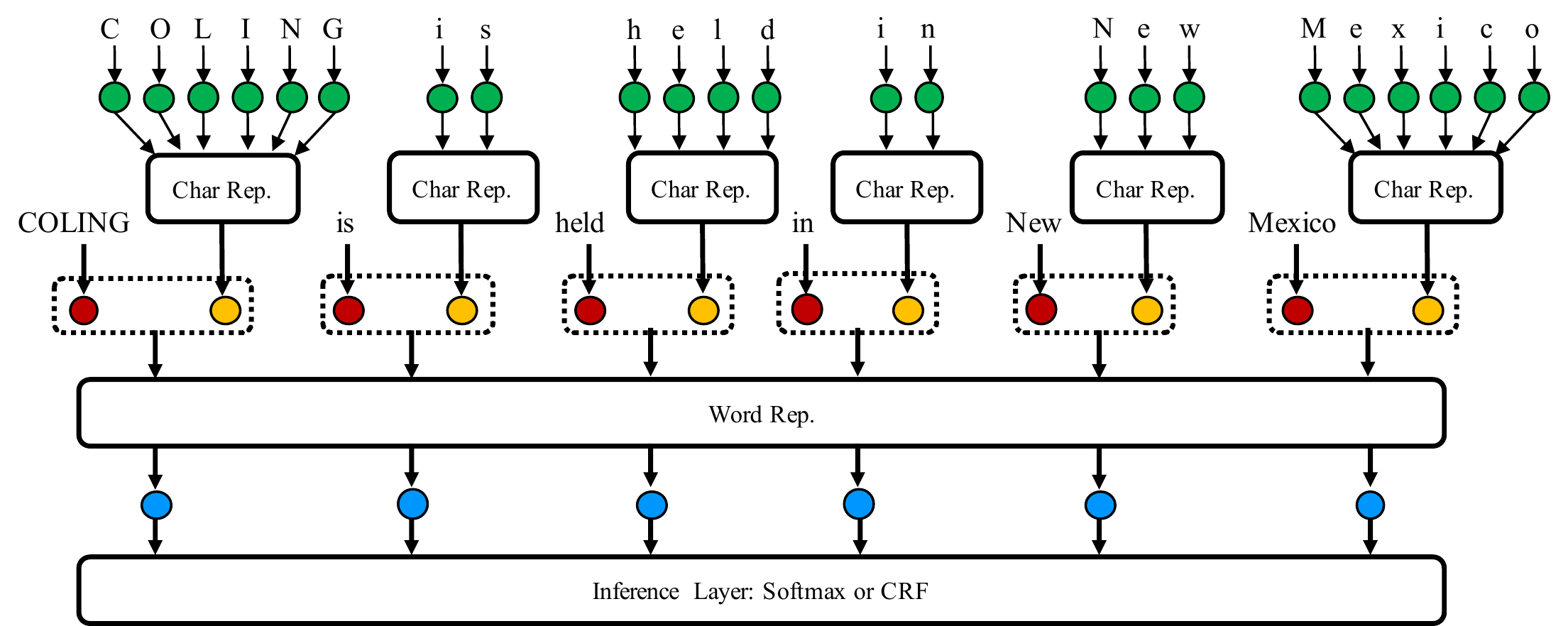}
  \caption{Neural sequence labeling architecture for sentence ``COLING is held in New Mexico''. Green, red, yellow and blue circles represent character embeddings, word embeddings, character sequence representations and word sequence representations, respectively.} 
  \label{fig:architecture} 
\end{figure}

\section{Related Work}
\newcite{collobert2011natural} proposed a seminal neural architecture for sequence labeling. It captures word sequence information with a one-layer CNN based on pretrained word embeddings and handcrafted neural features, followed with a CRF output layer. \newcite{dos2015boosting} extended this model by integrating character-level CNN features. \newcite{strubell2017fast} built a deeper dilated CNN architecture to capture larger local features. \newcite{hammerton2003named} was the first to exploit LSTM for sequence labeling. \newcite{huang2015bidirectional} built a BiLSTM-CRF structure, which has been extended by adding character-level LSTM \cite{lample2016neural,liu2017empower}, GRU \cite{yang2016multi}, and CNN \cite{chiu2015named,ma2016end} features. \newcite{yang2017neural} proposed a neural reranking model to improve NER models. These models achieve state-of-the-art results in the literature.



\newcite{reimers2017reporting} compared several word-based LSTM models for several sequence labeling tasks, reporting the score distributions over multiple runs rather than single value. They investigated the influence of various hyperparameters and configurations. Our work is similar in comparing different neural architectures under unified settings, but differs in four main aspects: 1) Their experiments are based on a BiLSTM with handcrafted word features, while our experiments are based on end-to-end neural models without human knowledge. 2) Their system gives relatively low performances on standard benchmarks\footnote{Based on their detailed experiment report \cite{reimers2017optimal}, the F1-scores on CoNLL 2003 NER task are generally less than 90\%, while \textit{state-of-the-art} results are around 91\%.}, while ours can give comparable or better results with state-of-the-art models, rendering our observations more informative for practitioners. 3) Our findings are more consistent with most previous work on configurations such as usefulness of character information \cite{lample2016neural,ma2016end}, optimizer \cite{chiu2015named,lample2016neural,ma2016end} and tag scheme \cite{ratinov2009design,dai2015enhancing}. In contrast, many results of \newcite{reimers2017reporting} contradict existing reports. 4) We conduct a wider range of comparison for word sequence representations, including all combinations of character CNN/LSTM and word CNN/LSTM structures, while \newcite{reimers2017reporting} studied the word LSTM models only.

\section{Neural Sequence Labeling Models}
Our neural sequence labeling framework contains three layers, i.e., a character sequence representation layer, a word sequence representation layer and an inference layer, as shown in Figure \ref{fig:architecture}. Character information has been proven to be critical for sequence labeling tasks \cite{chiu2015named,lample2016neural,ma2016end}, with LSTM and CNN being used to model character sequence information (``Char Rep.''). Similarly, on the word level,  LSTM or CNN structures can be leveraged to capture long-term information or local features (``Word Rep.''), respectively. Subsequently, the inference layer assigns labels to each word using the hidden states of word sequence representations.

\subsection{Character Sequence Representations}
Character features such as prefix, suffix and capitalization can be represented with embeddings through a feature-based lookup table \cite{collobert2011natural,huang2015bidirectional,strubell2017fast}, or neural networks without human-defined features \cite{lample2016neural,ma2016end}. In this work, we focus on neural character sequence representations without hand-engineered features.
\begin{figure} 
  \centering 
  \subfigure[Character CNN.]{ 
    \label{fig:charcnn} 
    \includegraphics[width=2.4in]{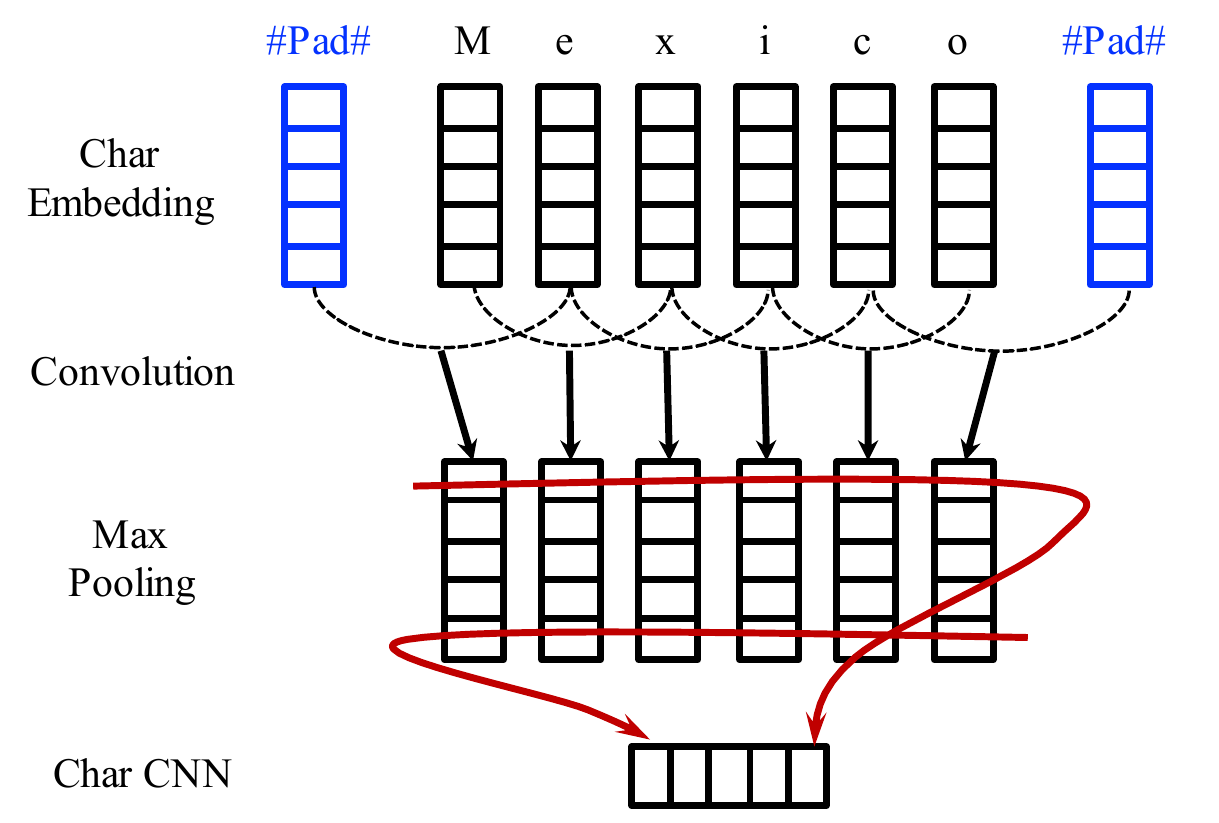}} 
  \subfigure[Character LSTM.]{ 
    \label{fig:charlstm} 
    \includegraphics[width=2.9in]{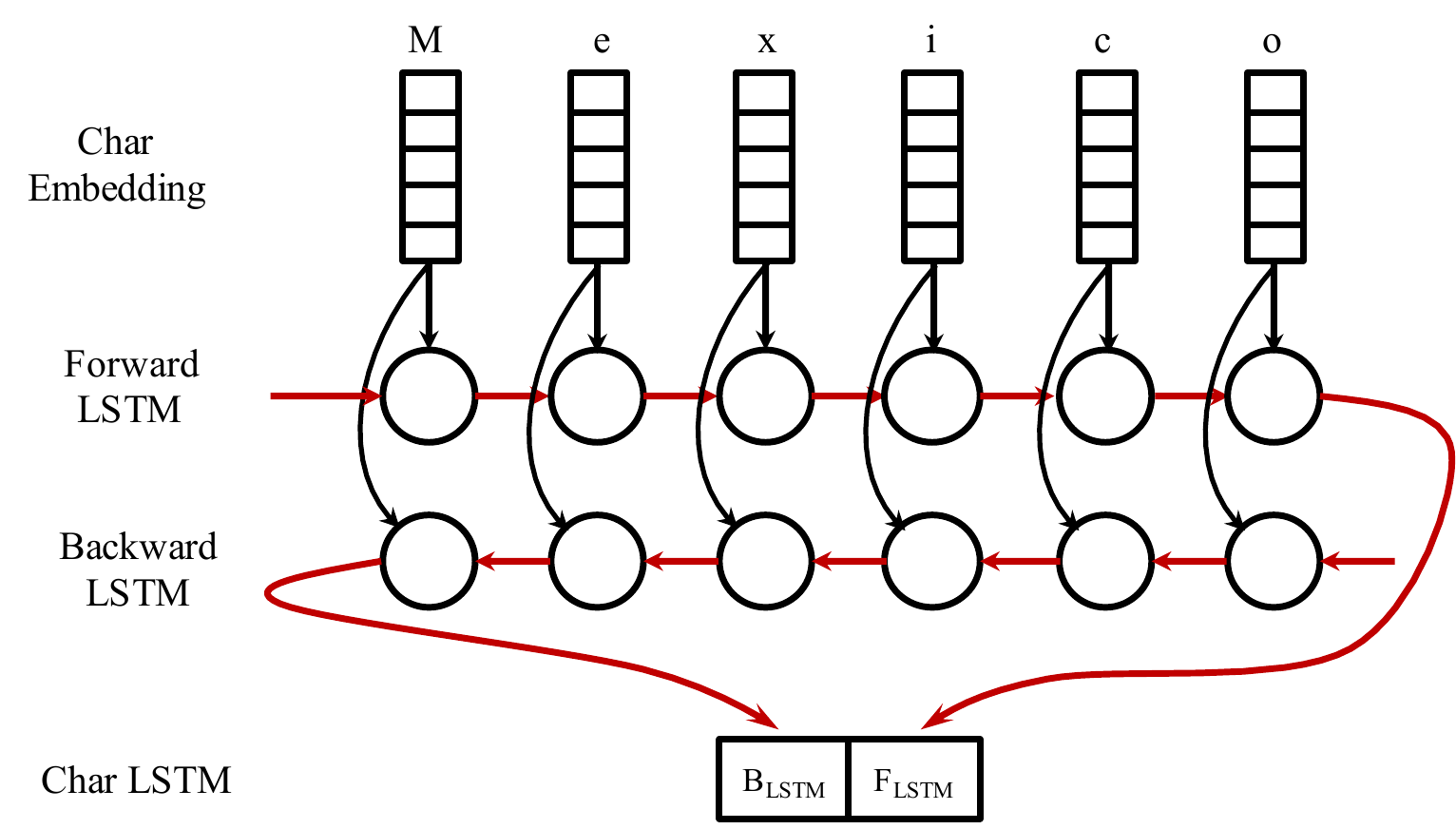}} 
  \caption{Neural character sequence representations.} 
  \label{fig:charfeature} 
\end{figure}

\textbf{Character CNN}. Using a CNN structure to encode character sequences was firstly proposed by \newcite{santos2014learning}, and followed by many subsequent investigations \cite{dos2015boosting,chiu2015named,ma2016end}. In our experiments, we take the same structure as \newcite{ma2016end}, using one layer CNN structure with \textit{max-pooling} to capture character-level representations. Figure \ref{fig:charcnn} shows the CNN structure on representing word ``Mexico''.

\textbf{Character LSTM}. Shown as Figure \ref{fig:charlstm}, in order to model the global character sequence information of a word ``Mexico'', we utilize a bidirectional LSTM on the character sequence of each word and concatenate the \textit{left-to-right} final state $F_{LSTM}$ and the \textit{right-to-left} final state $B_{LSTM}$ as character sequence representations. \newcite{liu2017empower} applied one bidirectional LSTM for the character sequence over a sentence rather than each word individually. We examined both structures and found that they give comparable accuracies on sequence labeling tasks. We choose \newcite{lample2016neural}'s structure as its character LSTMs can be calculated in parallel, making the system more efficient. 

\subsection{Word Sequence Representations}\label{ssec:wordrep}

Similar to character sequences in words, we can model word sequence information through LSTM or CNN structures. LSTM has been widely used in sequence labeling \cite{lample2016neural,ma2016end,chiu2015named,huang2015bidirectional,liu2017empower}. CNN can be much faster than LSTM due to the fact that convolution calculation can be parallel on the input sequence \cite{collobert2011natural,dos2015boosting,strubell2017fast}. 

\begin{figure} 
  \centering 
  \subfigure[Word CNN.]{ 
    \label{fig:wordcnn} 
    \includegraphics[width=2.52in]{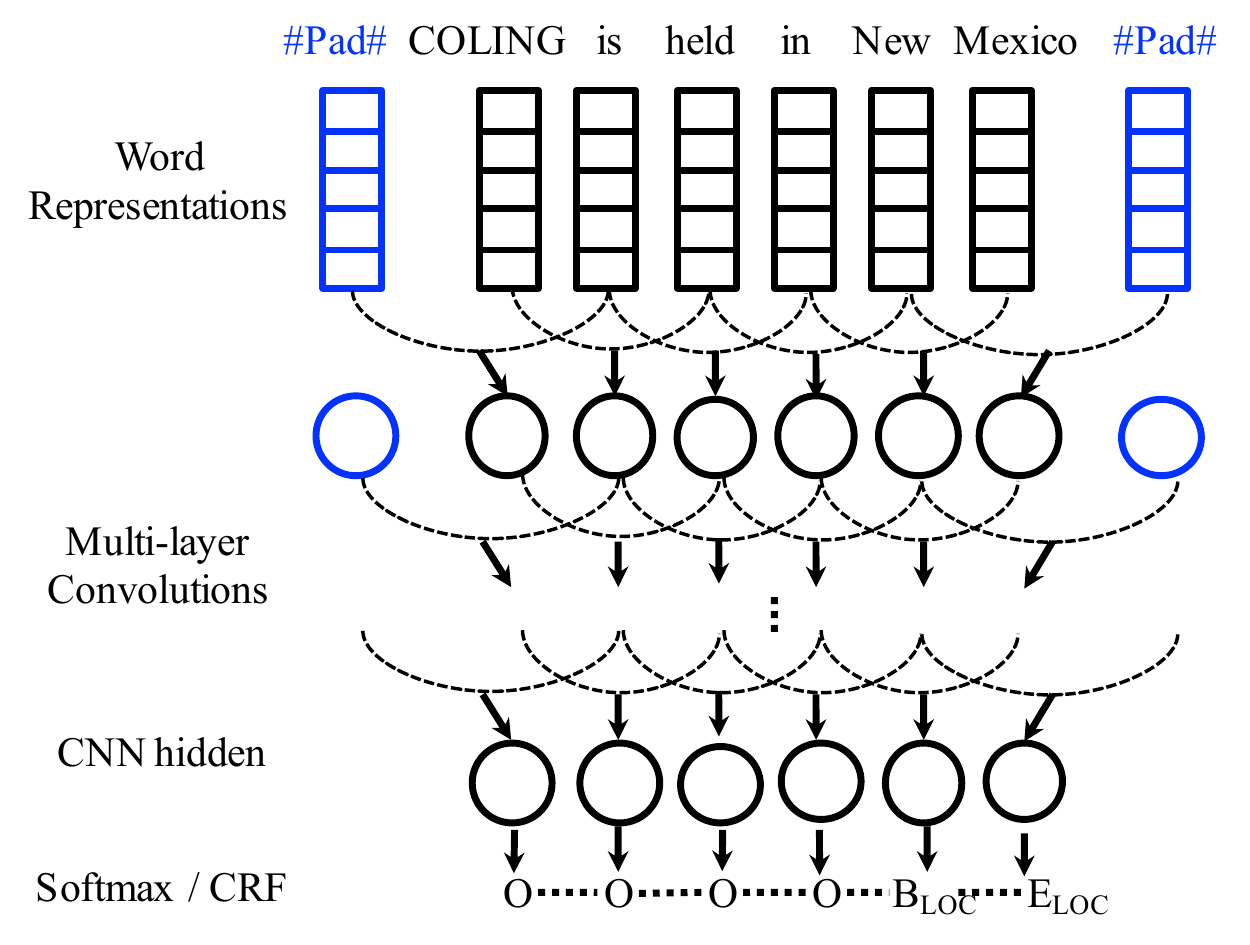}} 
  \subfigure[Word LSTM.]{ 
    \label{fig:wordlstm} 
    \includegraphics[width=2.68in]{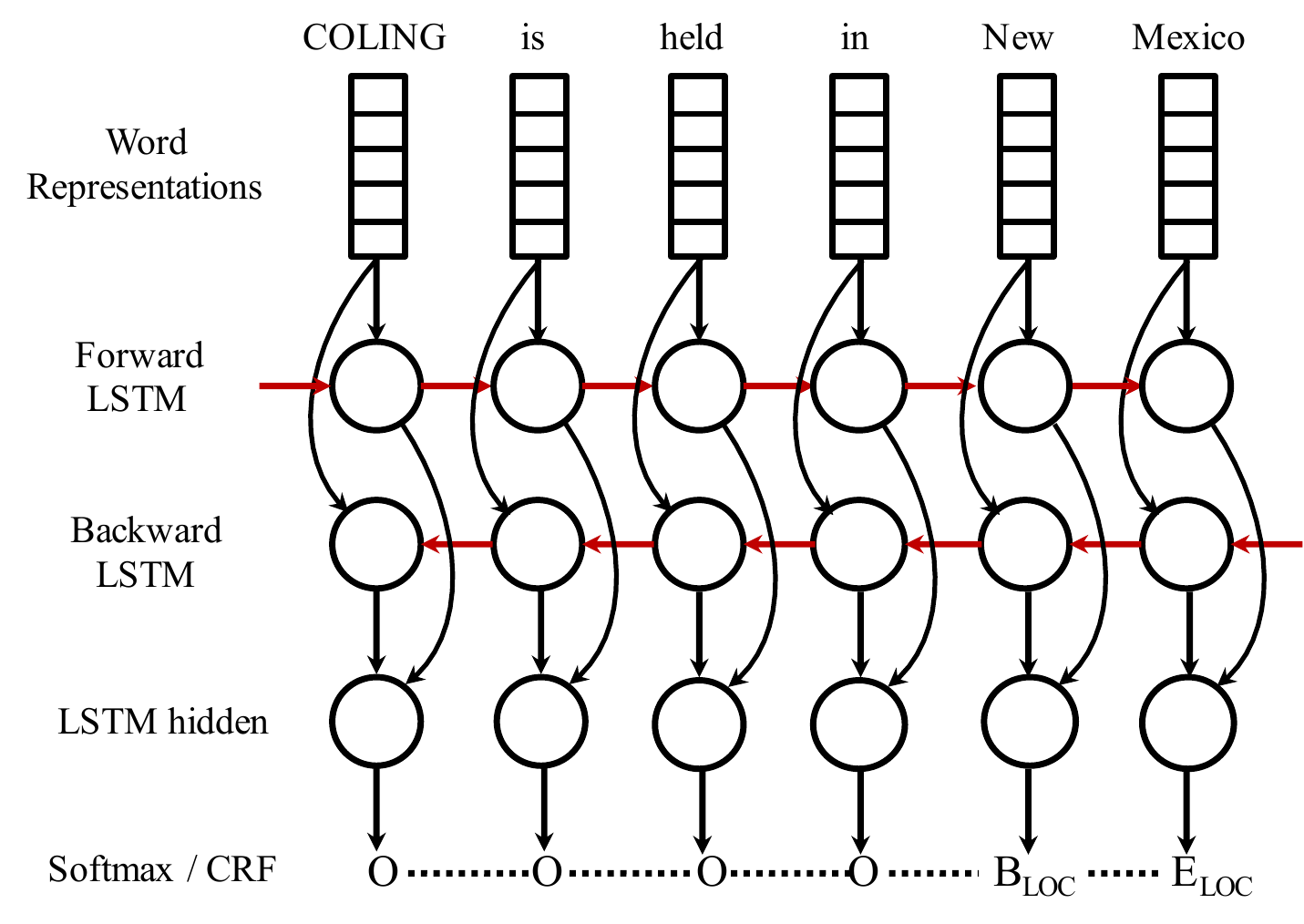}} 
  \caption{Neural word sequence representations.} 
  \label{fig:wordfeature} 
\end{figure}

\textbf{Word CNN}. Figure \ref{fig:wordcnn} shows the multi-layer CNN on word sequence, where words are represented by embeddings. If a character sequence representation layer is used, then word embeddings and character sequence representations are concatenated for word representations. For each CNN layer, a window of size 3 slides along the sequence, extracting local features on the word inputs and a ReLU function \cite{glorot2011deep} is followed. We follow \newcite{strubell2017fast} by using 4 CNN layers. Batch normalization \cite{ioffe2015batch} and dropout \cite{srivastava2014dropout} are applied following each CNN layer. 

\textbf{Word LSTM}. Shown in Figure \ref{fig:wordlstm}, word representations are fed into a forward LSTM and backward LSTM, respectively. The forward LSTM captures the word sequence information from left to right, while the backward LSTM extracts information in a reversed direction. The hidden states of the forward and backward LSTMs are concatenated at each word to give global information of the whole sequence.

\subsection{Inference Layer}\label{ssec:crf}
The inference layer takes the extracted word sequence representations as features and assigns labels to the word sequence. Independent local decoding with a linear layer mapping word sequence representations to label vocabulary and performing softmax can be quite effective \cite{ling2015finding}, while for tasks that with strong output label dependency, such as NER, CRF is a more appropriate choice. In this work, we examine both softmax and CRF as inference layer on three sequence labeling tasks.

\section{Experiments}
We investigate the main influencing factors to system accuracy, including the character sequence representations, word sequence representations, inference algorithm, pretrained embeddings, tag scheme, running environment and optimizer; analyzing system performances from the perspective of decoding speed and accuracies on in-vocabulary (IV) and out-of-vocabulary (OOV) entities/chunks/words.

\subsection{Settings}
\textbf{Data}. The NER dataset has been standardly split in \newcite{tjong2003introduction}. It contains four named entity types: \textsc{Person, Location, Organization,} and \textsc{Misc}. The chunking task is evaluated on CoNLL 2000 shared task \cite{tjong2000introduction}. We follow \newcite{reimers2017optimal} by using sections 15-18 of Wall Street Journal (WSJ) for training, section 19 as development set and section 20 as test set. For the POS tagging task, we use the WSJ portion of Peen Treebank, which has 45 POS tags. Following previous works \cite{toutanova2003feature,santos2014learning,ma2016end,liu2017empower}, we adopt the standard splits by using sections 0-18 as training set, sections 19-21 as development set and sections 22-24 as test set. No preprocessing is performed on either dataset except for normalizing digits. The dataset statistics are listed in Table \ref{tab:statistics}.

\textbf{Hyperparameters}. Table \ref{tab:hyperparameter} shows the hyperparameters used in our experiments, which mostly follow \newcite{ma2016end}, including the learning rate $\eta=0.015$ for word LSTM models. For word CNN based models, a large $\eta$ leads to convergence problem. We take $\eta=0.005$ with more epochs (200) instead. GloVe 100-dimension \cite{pennington2014glove} is used to initialize word embeddings and character embeddings are randomly initialized. We use mini-batch stochastic gradient descent (SGD) with a decayed learning rate to update parameters. For NER and chunking, we the BIOES tag scheme.

\textbf{Evaluation}. Standard precision (P), recall (R) and F1-score (F) are used as the evaluation metrics for NER and chunking; token accuracy is used to evaluate the performance of POS tagger. Development datasets are used to select the optimal model among all epochs, and we report scores of the selected model on the test dataset. To reduce the volatility of the system, we conduct each experiment 5 times under different random seeds, and report the \textit{max, mean,} and \textit{standard deviation} for each model.

\begin{table}
\parbox{.5\linewidth}{
\scalebox{0.8}{
\centering
\begin{tabular}{|c|l|c|c|c|c|}
            \hline
             \multicolumn{2}{|c|}{\textbf{Dataset}}& \textbf{Train}&\textbf{Dev} &\textbf{Test} & \textbf{Label}\\
            \hline
            \multirow{3}*{NER} & \#Sent &14,987   &3,644  &3,486   &\multirow{3}*{17}   \\
            & \#Token &205k   &52k   &47k   &   \\
            & \#Entity &23,523   &5,943   &5,654   &  \\
            \hline
            \multirow{3}*{chunking} & \#Sent &8,936   &1,844  & 2,012  &\multirow{3}*{42}   \\
            & \#Token & 212k  &44k   &47k   &   \\
            & \#Chunk &107k   &22k   &24k   &  \\
            \hline
            \multirow{2}*{POS} & \#Sent &38,219   &5,527  &5,462   &\multirow{2}*{45}    \\
            & \#Token &912k   &132k   &130k   &   \\
            \hline
  \end{tabular}
}
\caption{Statistics of datesets.}
\label{tab:statistics}
}
\hfill
\parbox{.5\linewidth}{
\scalebox{0.8}{
\centering
  \begin{tabular}{|l|l||l|l|}
    \hline \textbf{Parameter} &  \textbf{Value} & \textbf{Parameter} & \textbf{Value}\\ \hline
    char emb size & 30 & word emb size & 100 \\
    char hidden & 50 &word hidden&200    \\
    CNN window & 3 &word CNN layer &4\\
    batch size & 10  & dropout rate & 0.5\\
    $L_2$ regularization $\lambda$ & 1e-8 &learning rate decay & 0.05\\
    word LSTM $\eta$ & 0.015 &word CNN $\eta$ & 0.005\\
    word LSTM epochs & 100 &word CNN epochs & 200\\
    \hline
  \end{tabular}
}
\caption{Hyperparameters.}
\label{tab:hyperparameter}
}
\end{table}

\subsection{Results}
Tables \ref{tab:nerresult}, \ref{tab:chunkresult} and \ref{tab:posresult} show the results of the twelve models on NER, chunking and POS datasets, respectively. Existing work has also been listed in the tables for comparison. To simplify the description, we use ``CLSTM'' and ``CCNN'' to represent character LSTM and character CNN encoder, respectively. Similarly, ``WLSTM'' and ``WCNN'' represent word LSTM and word CNN structure, respectively.

As shown in Table \ref{tab:nerresult}, most NER work focuses on WLSTM+CRF structures with different character sequence representations. We re-implement the structure of several reports \cite{chiu2015named,ma2016end,peters2017semi}, which take the CCNN+WLSTM+CRF architecture. Our reproduced models give slightly better performances. The results of \newcite{lample2016neural} can be reproduced by our CLSTM+WLSTM+CRF. In most cases, our ``Nochar'' based models underperform their corresponding prototypes \cite{huang2015bidirectional,strubell2017fast}, which utilize the hand-crafted features. 

Table \ref{tab:chunkresult} shows the results of the chunking task. \newcite{peters2017semi} give the best reported single model results (95.00$\pm$0.08), and our CLSTM+WLSTM+CRF gives a comparable performance (94.93$\pm$0.05). We re-implement \newcite{zhai2017neural}'s model in our Nochar+WLSTM but cannot reproduce their results, this may because that they use grid search for hyperparameter selection. Our Nochar+WCNN+CRF can give comparable results with \newcite{collobert2011natural}, even ours does not include character information.

The results of the POS tagging task is shown in Table \ref{tab:posresult}. The results of \newcite{lample2016neural}, \newcite{ma2016end} and \newcite{yang2017transfer} can be reproduced by our CLSTM+WLSTM+CRF and CCNN+WLSTM+CRF models. Our WLSTM based models give better results than WLSTM+CRF based models, this is consistent with the fact that \newcite{ling2015finding} take CLSTM+WLSTM without CRF layer but achieve the best POS accuracy. \newcite{santos2014learning} build a pure CNN structure on both character and word level, which can be reproduced by our CCNN+WCNN models. 

Based on above observations, most results in the literature are reproducible. Our implementations can achieve the comparable or better results with state-of-the-art models. We do not fine-tune any hyperparameter to fit the specific task. Results on Table \ref{tab:nerresult}, \ref{tab:chunkresult} and \ref{tab:posresult} are all under the same hyperparameters, which demonstrates the generalization ability of our framework.

\subsection{Network settings}\label{ssc:archi}
\textbf{Character LSTM vs Character CNN}. Unlike the observations of \newcite{reimers2017reporting}, in our experiments, character information can significantly ($p<0.01$)\footnote{We use \textit{t-test} to calculate the $p$ value, reporting the highest $p$ value when giving the conclusions on multiple configurations.} improve sequence labeling models (by comparing the row of Nochar with CLSTM or CCNN on Table \ref{tab:nerresult}, \ref{tab:chunkresult} and \ref{tab:posresult}), while the difference between CLSTM and CCNN is not significant. In most cases, CLSTM and CCNN can give comparable results under different frameworks and different tasks. CCNN gives the best NER result under the WLSTM+CRF framework, while CLSTM gets better NER results in all other configurations. For chunking and POS tagging, CLSTM consistently outperforms CCNN under all settings, while the difference is statistically insignificant ($p>0.2$). We conclude that the difference between CLSTM and CCNN is small, which is consistent with the observation of \newcite{reimers2017reporting}.

\textbf{Word LSTM vs Word CNN}. By comparing the performances of WLSTM+CRF, WLSTM with WCNN+CRF, WCNN on the three benchmarks, we conclude that word-based LSTM models are significantly ($p<0.01$) better than word-based CNN models in most cases. It demonstrates that global word context information is necessary for sequence labeling.

\textbf{Softmax vs CRF}. Models with CRF inference layer can consistently outperform the models with softmax layer under all configurations on NER and chunking tasks, proving the effectiveness of label dependency information. While for POS tagging, the local softmax based models give slightly better accuracies while the difference is insignificant ($p>0.2$).

\begin{table}[!tp]
\begin{center}
\resizebox{.86\columnwidth}{!}{%
\begin{tabular}{|l|ll|l|l|l|l|}
\hline 
\multicolumn{3}{|c|}{\multirow{2}*{\textbf{Results (F1-score)}}}&\multicolumn{4}{|c|}{\textbf{NER}}\\
\cline{4-7}
\multicolumn{3}{|c|}{\textbf{}}&\textbf{WLSTM+CRF}& \textbf{WLSTM} &\textbf{WCNN+CRF} &\textbf{WCNN} \\ 
\hline
\hline
\multirow{5}*{\textbf{Nochar}}
&\multicolumn{2}{|c|}{\multirow{3}*{Literature}} &90.10 (H-15)* &87.00 (M-16) &89.59 (C-11)* &89.97 (S-17)*\\
&&  &90.20 (L-16) &89.34 (S-17)* &90.54 (S-17)* &\\
&&  &90.43 (S-17)* & &  & \\
\cline{2-7}
&\multicolumn{1}{l|}{\multirow{2}{*}{Ours}} &Max  &89.45 &88.57 &88.90 &88.56\\
&\multicolumn{1}{l|}{} &Mean$\pm$std &89.31$\pm$0.10  &88.49$\pm$0.17 &88.65$\pm$0.20&88.50$\pm$0.05\\
\hline
\hline
\multirow{4}*{\textbf{CLSTM}}
&\multicolumn{2}{|c|}{\multirow{2}*{Literature}} &90.94 (L-16) &89.15 (L-16) &\multirow{2}*{--}  &\multirow{2}*{--} \\
&&  &91.20 (Y-17)\ddag &  &  & \\
\cline{2-7}
&\multicolumn{1}{l|}{\multirow{2}{*}{Ours}}  &Max &91.20  &90.84 &90.70 &90.46\\
&\multicolumn{1}{l|}{} &Mean$\pm$std &91.08$\pm$0.08  &90.77$\pm$0.06 &90.48$\pm$0.23 &90.28$\pm$0.30\\
\hline
\hline
\multirow{5}*{\textbf{CCNN}}
&\multicolumn{2}{|c|}{\multirow{3}*{Literature}}  &90.91$\pm$0.20 (C-16) &89.36 (M-16) &\multirow{3}*{--}  &\multirow{3}*{--} \\
&&  &91.21 (M-16) &  &  & \\
&&  &90.87$\pm0.13$ (P-17) &  &  & \\
\cline{2-7}
&\multicolumn{1}{l|}{\multirow{2}{*}{Ours}} &Max  &91.35  &90.73 &90.43 &90.51\\
&\multicolumn{1}{l|}{} &Mean$\pm$std  &91.11$\pm$0.21  &90.60$\pm$0.11  &90.28$\pm$0.09 &90.26$\pm$0.19\\
\hline
\end{tabular}
}
\end{center}
\caption{Results for NER.\footnote{\label{note1} In Tables \ref{tab:nerresult}, \ref{tab:chunkresult} and \ref{tab:posresult}, the abbreviation (C-11)=\newcite{collobert2011natural}, (S-14)=\newcite{santos2014learning}, (H-15)=\newcite{huang2015bidirectional}, (L-16)=\newcite{lample2016neural}, (M-16)=\newcite{ma2016end}, (C-16)=\newcite{chiu2015named}, (Z-17)=\newcite{zhai2017neural}, (H-17)=\newcite{hashimoto2017joint}, (Y-17)=\newcite{yang2017transfer}, (R-17)=\newcite{rei2017semi}, (S-17)=\newcite{strubell2017fast} and (P-17)=\newcite{peters2017semi}. * suggests that models with handcrafted features. Results of (L-16)\dag $\;$ is reported by \newcite{liu2017empower} by running the code of \newcite{lample2016neural} on the corresponding dataset. (Y-17)\ddag $\;$ use GRU for character and word sequence representations; here we regard GRU as a variant of LSTM.}}
\label{tab:nerresult}
\end{table}

\begin{table}[!tp]
\begin{center}
\resizebox{.86\columnwidth}{!}{%
\begin{tabular}{|l|ll|l|l|l|l|}
\hline 
\multicolumn{3}{|c|}{\multirow{2}*{\textbf{Results (F1-score)}}}&\multicolumn{4}{|c|}{\textbf{chunking}}\\
\cline{4-7}
\multicolumn{3}{|c|}{\textbf{}}&\textbf{WLSTM+CRF}& \textbf{WLSTM} &\textbf{WCNN+CRF} &\textbf{WCNN} \\ 
\hline
\hline
\multirow{4}*{\textbf{Nochar}}
&\multicolumn{2}{|c|}{\multirow{2}*{Literature}}   &94.46 (H-15)* &94.13 (Z-17) &94.32 (C-11)* &\multirow{2}*{--}\\
&& &  &95.02 (H-17)* & &\\
\cline{2-7}
&\multicolumn{1}{l|}{\multirow{2}{*}{Ours}} &Max  &94.49&93.79&94.23&94.12\\
&\multicolumn{1}{l|}{}&Mean$\pm$std &94.37$\pm$0.11&93.75$\pm$0.04&94.11$\pm$0.08&94.08$\pm$0.06\\
\hline
\hline
\multirow{4}*{\textbf{CLSTM}}
&\multicolumn{2}{|c|}{\multirow{2}*{Literature}}  &93.15 (R-17) &\multirow{2}*{--} &\multirow{2}*{--}  &\multirow{2}*{--} \\
&&  &94.66 (Y-17)\ddag & &  & \\
\cline{2-7}
&\multicolumn{1}{l|}{\multirow{2}{*}{Ours}}&Max &95.00&94.33&94.76&94.55\\
&\multicolumn{1}{l|}{}&Mean$\pm$std &94.93$\pm$0.05&94.28$\pm$0.04&94.66$\pm$0.01&94.48$\pm$0.07\\
\hline
\hline
\multirow{3}*{\textbf{CCNN}}
&\multicolumn{2}{|c|}{\multirow{1}*{Literature}}  &95.00$\pm$0.08 (P-17) &\multirow{1}*{--}  &\multirow{1}*{--}  &\multirow{1}*{--} \\
\cline{2-7}
&\multicolumn{1}{l|}{\multirow{2}{*}{Ours}} &Max  &95.06&94.24&94.77&94.51\\
&\multicolumn{1}{l|}{}&Mean$\pm$std  &94.86$\pm$0.14&94.19$\pm$0.04&94.66$\pm$0.13&94.47$\pm$0.03\\
\hline
\end{tabular}
}
\end{center}
\caption{Results for chunking.$^4$}
\label{tab:chunkresult}
\end{table}

\begin{table}[!tp]
\begin{center}
\resizebox{.86\columnwidth}{!}{%
\begin{tabular}{|l|ll|l|l|l|l|}
\hline 
\multicolumn{3}{|c|}{\multirow{2}*{\textbf{Results (Accuracy)}}}&\multicolumn{4}{|c|}{\textbf{POS}}\\
\cline{4-7}
\multicolumn{3}{|c|}{\textbf{}}&\textbf{WLSTM+CRF}& \textbf{WLSTM} &\textbf{WCNN+CRF} &\textbf{WCNN} \\ 
\hline
\hline
\multirow{4}*{\textbf{Nochar}}
&\multicolumn{2}{|c|}{\multirow{2}*{Literature}}   &97.55 (H-15)* &96.93 (M-16) &97.29 (C-11)* &96.13 (S-14)\\
&&&&97.45 (H-17)*&&\\
\cline{2-7}
&\multicolumn{1}{l|}{\multirow{2}{*}{Ours}}&Max  &97.20&97.23&96.99&97.07\\
&\multicolumn{1}{l|}{}&Mean$\pm$std &97.19$\pm$0.01&97.20$\pm$0.02&96.95$\pm$0.04&97.01$\pm$0.04\\
\hline
\hline
\multirow{4}*{\textbf{CLSTM}}
&\multicolumn{2}{|c|}{\multirow{2}*{Literature}}  &97.35$\pm$0.09 (L-16)\dag &97.78 (L-15) &\multirow{2}*{--}  &\multirow{2}*{--} \\
&&  &97.55 (Y-17)\ddag &  &  & \\
\cline{2-7}
&\multicolumn{1}{l|}{\multirow{2}{*}{Ours}}&Max &97.49&97.51&97.38&97.38\\
&\multicolumn{1}{l|}{}&Mean$\pm$std &97.47$\pm$0.02&97.48$\pm$0.02&97.33$\pm$0.03&97.33$\pm$0.04\\
\hline
\hline
\multirow{3}*{\textbf{CCNN}}
&\multicolumn{2}{|c|}{\multirow{1}*{Literature}}   &97.55 (M-16) &97.33 (M-16) &\multirow{1}*{--}  &97.32 (S-14) \\
\cline{2-7}
&\multicolumn{1}{l|}{\multirow{2}{*}{Ours}}&Max  &97.46&97.51&97.33&97.33\\
&\multicolumn{1}{l|}{}&Mean$\pm$std  &97.43$\pm$0.02&97.44$\pm$0.04&97.29$\pm$0.03&97.30$\pm$0.02\\
\hline
\end{tabular}
}
\end{center}
\caption{Results for POS tagging.$^4$}
\label{tab:posresult}
\end{table}

\begin{figure} 
  \centering 
  \subfigure[Pretrained embeddings.]{ 
    \label{fig:embcompare} 
    \includegraphics[width=3in]{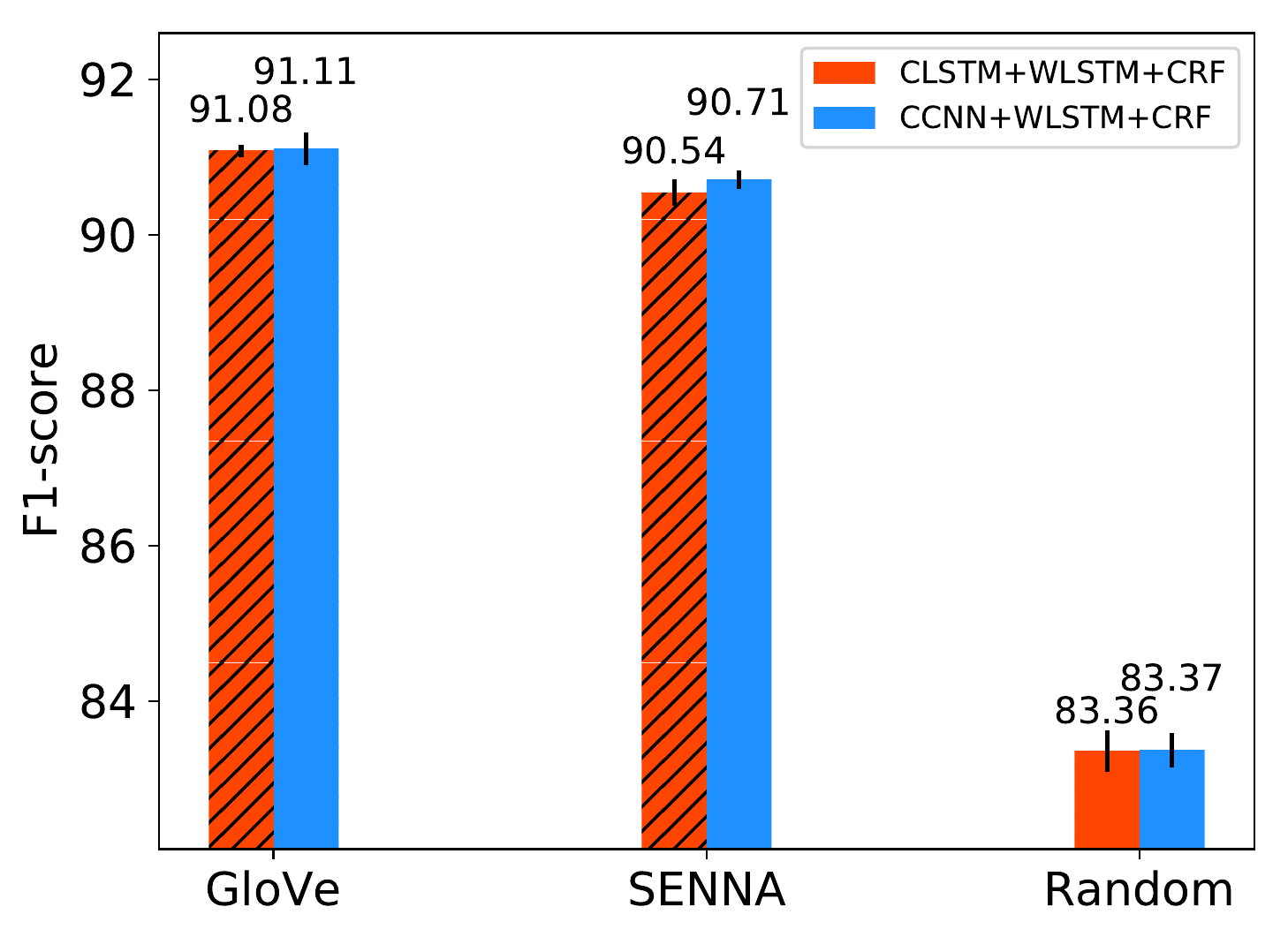}} 
  \subfigure[Tag scheme and running environment.]{ 
    \label{fig:allsetting} 
    \includegraphics[width=3.1in]{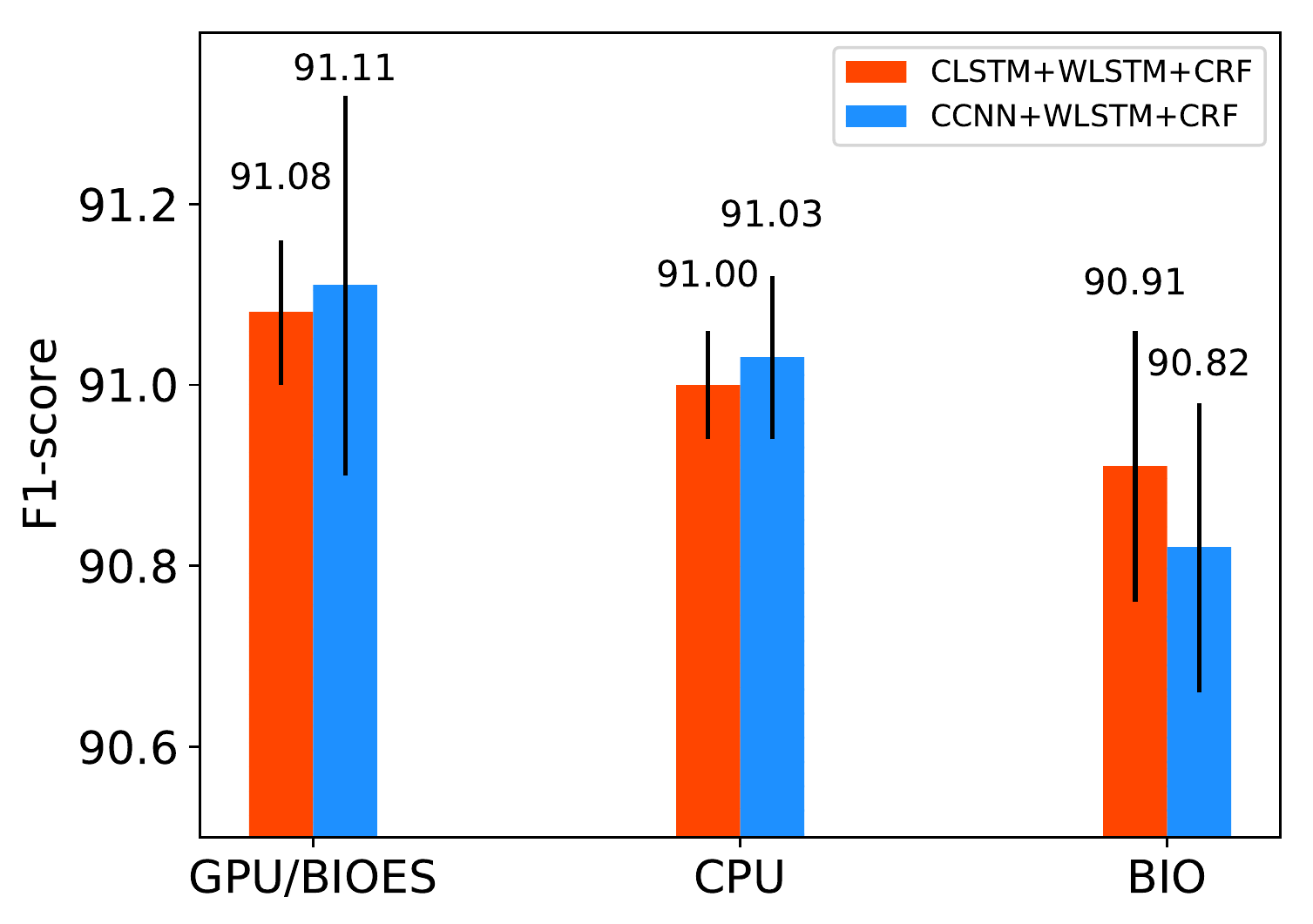}} 
  \caption{Performance comparison on the CoNLL 2003 NER task.} 
  \label{fig:setting} 
\end{figure}

\subsection{External factors} \label{sc:influencefactor}
In addition to model structures, external factors such as pretrained embeddings, tag scheme, and optimizer can significantly influence system performance. We investigate a set of external factors on the NER dataset with the two best models: CLSTM+WLSTM+CRF and CCNN+WLSTM+CRF.

\textbf{Pretrained embedding}. Figure \ref{fig:embcompare} shows the F1-scores of the two best models on the NER test set with two different pretrained embeddings, as well as the random initialization. Compared with the random initialization, models using pretrained embeddings give significant improvements ($p<0.01$). The GloVe 100-dimension embeddings give higher F1-scores than SENNA \cite{collobert2011natural} on both models, which is consistent with the observation of \newcite{ma2016end}.

\textbf{Tag scheme}. We examine two different tag schemes: BIO and BIOES \cite{ratinov2009design}. The results are shown in Figure \ref{fig:allsetting}. In our experiments, models using BIOES are significantly ($p<0.05$) better than BIO. Our observation is consistent with most literature \cite{ratinov2009design,dai2015enhancing}. \newcite{reimers2017reporting} report that the difference between the schemes is insignificant.

\textbf{Running environment}. \newcite{liu2017empower} observe that neural sequence labeling models can give better results on GPU rather than CPU. We conduct repeated experiments on both GPU and CPU environments. The results are shown in Figure \ref{fig:allsetting}. Models run on CPU give a lower mean F1-score than models run on GPU, while the difference is insignificant ($p>0.2$).

\textbf{Optimizer}. We compare different optimizers including SGD, Adagrad \cite{duchi2011adaptive}, Adadelta \cite{zeiler2012adadelta}, RMSProp \cite{tieleman2012lecture} and Adam \cite{kingma2014adam}. The results are shown in Figure \ref{fig:optisetting}\footnote{We fine-tune the learning rates for Adagrad, Adadelta, RMSProp and Adam, and report the best results in the figure.}. In contrast to \newcite{reimers2017reporting}, who reported that SGD is the worst optimizer, our results show that SGD outperforms all other optimizers significantly ($p<0.01$), with a slower convergence process during training. Our observation is consistent with most literature \cite{chiu2015named,lample2016neural,ma2016end}.

\subsection{Analysis}

\begin{figure}
\centering
\begin{minipage}{.5\textwidth}
  \centering
  \includegraphics[width=3.1in]{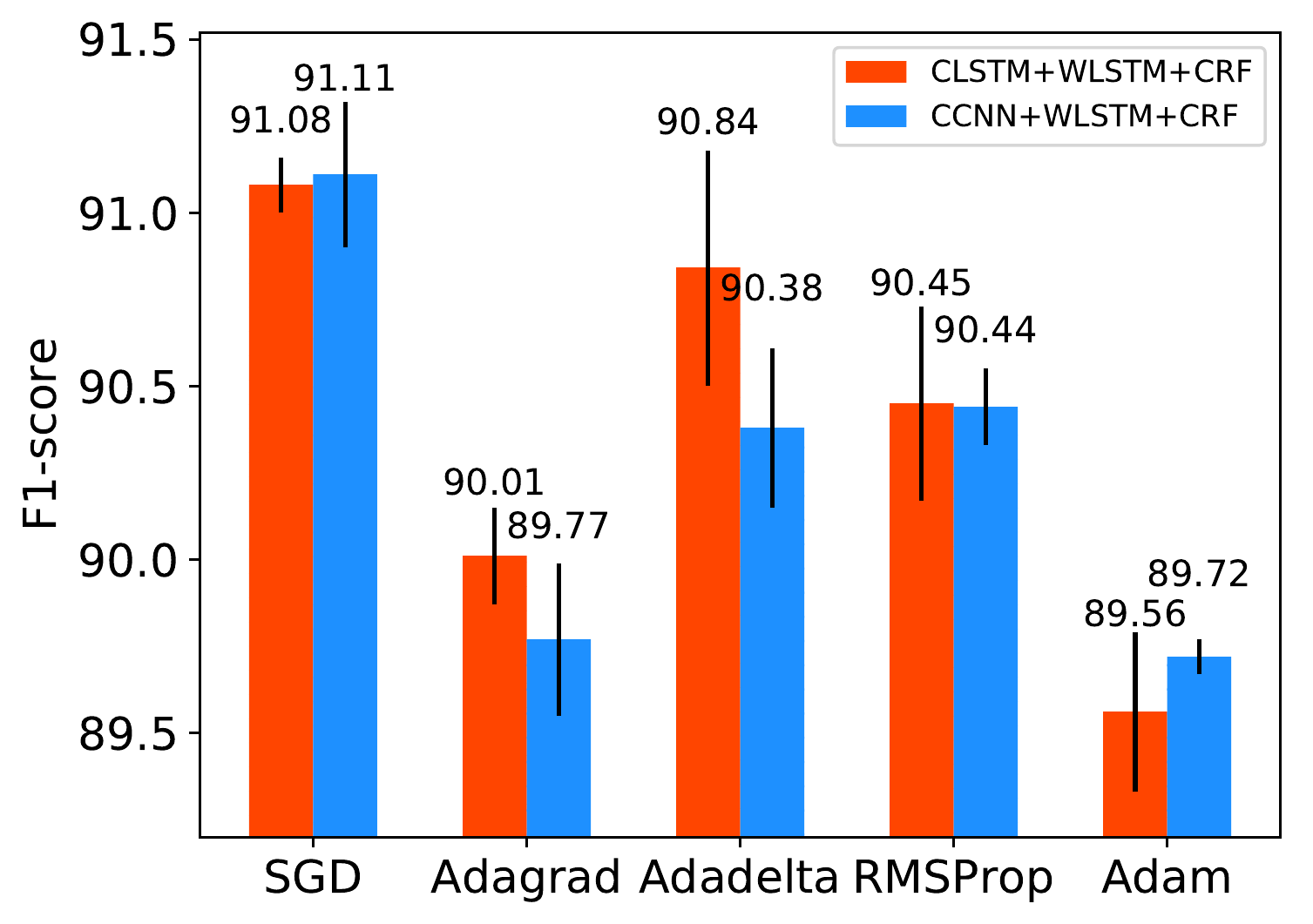}
  \caption{Optimizers.}
  \label{fig:optisetting}
\end{minipage}%
\begin{minipage}{.5\textwidth}
  \centering
  \includegraphics[width=3in]{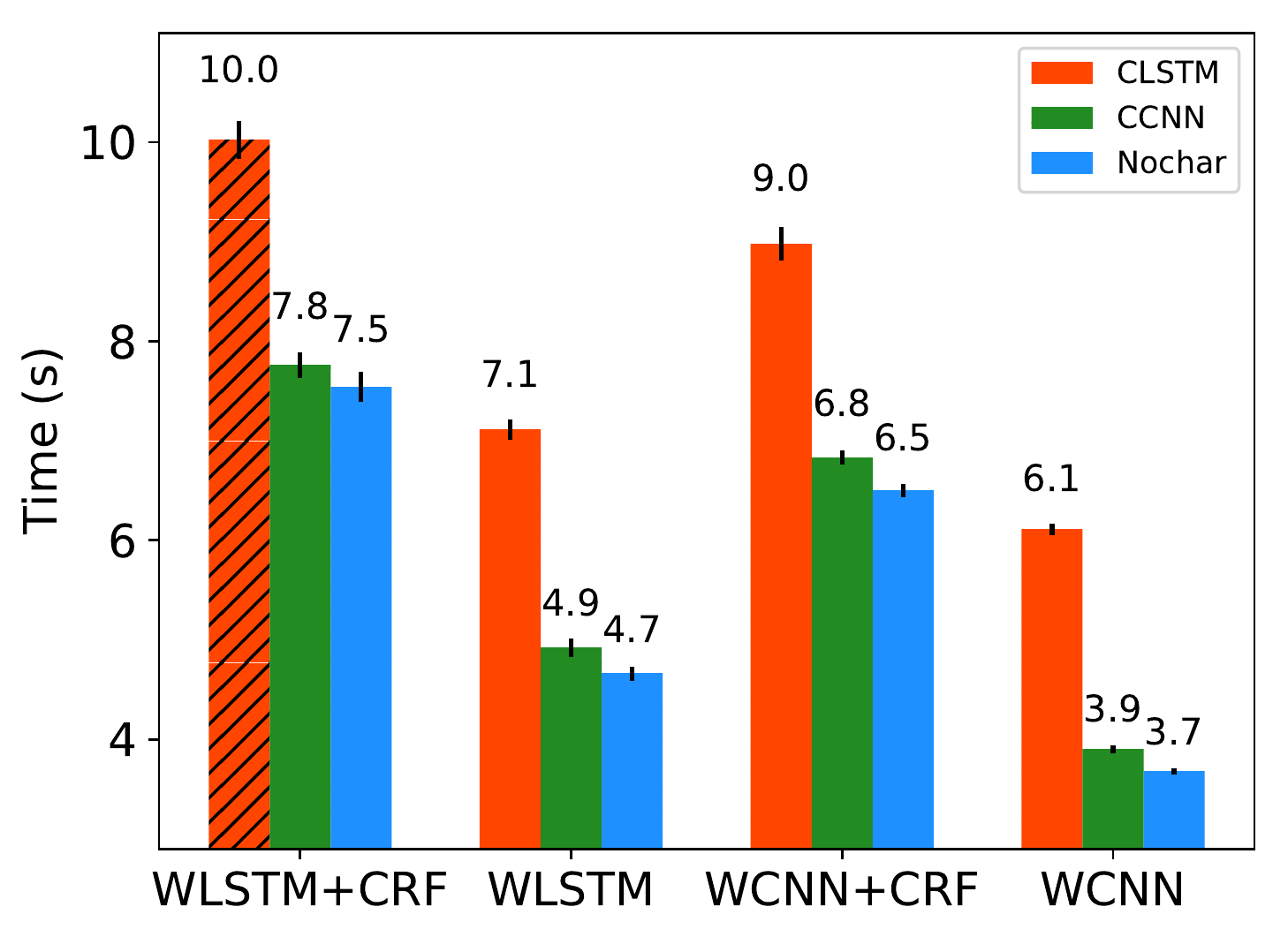}
  \caption{Decoding times (10000 NER sentences).}
  \label{fig:decodespeed}
\end{minipage}
\end{figure}

\textbf{Decoding speed}. We test the decoding speeds of the twelve models on the NER dataset using a Nvidia GTX 1080 GPU. Figure \ref{fig:decodespeed} shows the decoding times on 10000 NER sentences. The CRF inference layer severely limits the decoding speed due to the left-to-right inference process, which disables the parallel decoding. Character LSTM significantly slows down the system. Compared with models without character information, adding character CNN representations does not affect the decoding speed too much but can give significant accuracy improvements (shown in Section \ref{ssc:archi}). Due to the support of parallel computing, word-based CNN models are consistently faster than word-based LSTM models, with close accuracies, leading to large utilization potential in practice.

\textbf{OOV error}. We conduct error analysis on in-vocabulary and out-of-vocabulary words with the CRF based models\footnote{We choose the models that give the median performance on the test set for conducting result analysis.}. Following \newcite{ma2016end}, words in the test set are divided into four subsets: in-vocabulary words, out-of-training-vocabulary words (OOTV), out-of-embedding-vocabulary words (OOEV) and out-of-both-vocabulary words (OOBV). For NER and chunking, we consider entities or chunks rather than words. The OOV entities and chunks are categorized following \newcite{ma2016end}.

Table \ref{tab:OOV} shows the performances of different OOV splits on three benchmarks. The top three rows list the performances of word-based LSTM CRF models, followed by the word-based CNN CRF models. The results of OOEV in NER keep 100\% because of there exist only 8 OOEV entities and all are recognized correctly. It is obvious that character LSTM or CNN representations improve OOTV and OOBV the most on both WLSTM+CRF and WCNN+CRF models across all three datasets, proving that the main contribution of neural character sequence representations is to disambiguate the OOV words. Models with character LSTM representations give the best IV scores across all configurations, which may be because character LSTM can be well trained on IV data, bringing the useful global character sequence information. On the OOVs, character LSTM and CNN gives comparable results.

\begin{table}[!tp]
\begin{center}
\resizebox{\columnwidth}{!}{%
\begin{tabular}{|l|llll|llll|llll|}
\hline 
\multirow{2}*{\textbf{Results}} &\multicolumn{4}{|c|}{\textbf{NER (F1-score)}} &\multicolumn{4}{|c|}{\textbf{chunking (F1-score)}} &\multicolumn{4}{|c|}{\textbf{POS (Accuracy)}}\\
\cline{2-13}
 &IV &OOTV &OOEV &OOBV &IV &OOTV &OOEV &OOBV &IV &OOTV &OOEV &OOBV\\ 
\hline
Nochar+WLSTM+CRF &91.33 &87.36 &100.00 &69.68 &94.87 &90.84 &95.51 &91.47 &97.51 &89.76 &94.07 &75.36\\ 
\hline
CLSTM+WLSTM+CRF &\textbf{92.18} &90.63 &100.00 &78.57 &\textbf{95.20} &\textbf{92.65} &94.38 &\textbf{94.01} &\textbf{97.63} &\textbf{93.82} &94.07 &\textbf{87.32}\\ 
\hline
CCNN+WLSTM+CRF &91.76 &\textbf{91.25} &100.00 &\textbf{81.58} &95.15 &92.34 &\textbf{97.75} &93.55 &97.62 &93.33 &\textbf{94.69} &83.82\\ 
\hline
\hline
Nochar+WCNN+CRF &90.71 &86.99 &100.00 &69.09 &94.56 &90.98 &93.26 &91.71 &97.29 &89.10 &\textbf{94.17} &74.15\\ 
\hline
CLSTM+WCNN+CRF &\textbf{91.59} &90.07 &100.00 &77.92 &\textbf{95.02} &91.86 &94.38 &\textbf{93.32} &\textbf{97.48} &\textbf{93.28} &\textbf{94.17} &\textbf{88.29}\\ 
\hline
CCNN+WCNN+CRF &91.35 &\textbf{90.46} &100.00 &\textbf{78.88} &94.83 &\textbf{92.42} &\textbf{96.63} &92.40 &97.46 &92.74 &93.86 &87.80\\ 
\hline
\end{tabular}
}
\end{center}
\caption{Results for OOV analysis.}
\label{tab:OOV}
\end{table}

\section{Conclusion}
We built a unified neural sequence labeling framework to reproduce and compare recent state-of-the-art models with different configurations. We explored three neural model design decisions: character sequence representations, word sequence representations, and inference layer. Experiments show that character information helps to improve model performances, especially on disambiguating OOV words. Character-level LSTM and CNN structures give comparable improvements, with the latter being more efficient. In most cases, models with word-level LSTM encoders outperform those with CNN, at the expense of longer decoding time. We observed that the CRF inference algorithm is effective on NER and chunking tasks, but does not have the advantage on POS tagging. With controlled experiments on the NER dataset, we showed that BIOES tags are better than BIO. Besides, pretrained GloVe 100d embedding and SGD optimizer give significantly better performances compared to their competitors.

\section*{Acknowledgements}
We thank the anonymous reviewers for their useful comments. Yue Zhang is the corresponding author.


\bibliographystyle{acl}
\bibliography{coling2018}

\begin{thebibliography}{}

\bibitem[\protect\citename{Chiu and Nichols}2016]{chiu2015named}
Jason Chiu and Eric Nichols.
\newblock 2016.
\newblock Named entity recognition with bidirectional {LSTM-CNNs}.
\newblock {\em TACL}, 4:357--370.

\bibitem[\protect\citename{Collobert \bgroup et al.\egroup
  }2011]{collobert2011natural}
Ronan Collobert, Jason Weston, L{\'e}on Bottou, Michael Karlen, Koray
  Kavukcuoglu, and Pavel Kuksa.
\newblock 2011.
\newblock Natural language processing (almost) from scratch.
\newblock {\em Journal of Machine Learning Research}, 12(Aug):2493--2537.

\bibitem[\protect\citename{Dai \bgroup et al.\egroup }2015]{dai2015enhancing}
Hong-Jie Dai, Po-Ting Lai, Yung-Chun Chang, and Richard Tzong-Han Tsai.
\newblock 2015.
\newblock Enhancing of chemical compound and drug name recognition using
  representative tag scheme and fine-grained tokenization.
\newblock {\em Journal of cheminformatics}, 7(S1):S14.

\bibitem[\protect\citename{dos Santos \bgroup et al.\egroup
  }2015]{dos2015boosting}
C{\i}cero dos Santos, Victor Guimaraes, RJ~Niter{\'o}i, and Rio de~Janeiro.
\newblock 2015.
\newblock Boosting named entity recognition with neural character embeddings.
\newblock In {\em Proceedings of NEWS 2015 The Fifth Named Entities Workshop},
  page~25.

\bibitem[\protect\citename{Duchi \bgroup et al.\egroup
  }2011]{duchi2011adaptive}
John Duchi, Elad Hazan, and Yoram Singer.
\newblock 2011.
\newblock Adaptive subgradient methods for online learning and stochastic
  optimization.
\newblock {\em Journal of Machine Learning Research}, 12(Jul):2121--2159.

\bibitem[\protect\citename{Glorot \bgroup et al.\egroup }2011]{glorot2011deep}
Xavier Glorot, Antoine Bordes, and Yoshua Bengio.
\newblock 2011.
\newblock Deep sparse rectifier neural networks.
\newblock In {\em Proceedings of the Fourteenth International Conference on
  Artificial Intelligence and Statistics}, pages 315--323.

\bibitem[\protect\citename{Hammerton}2003]{hammerton2003named}
James Hammerton.
\newblock 2003.
\newblock Named entity recognition with long short-term memory.
\newblock In {\em Proceedings of the seventh conference on Natural language
  learning at HLT-NAACL 2003-Volume 4}, pages 172--175. Association for
  Computational Linguistics.

\bibitem[\protect\citename{Hashimoto \bgroup et al.\egroup
  }2017]{hashimoto2017joint}
Kazuma Hashimoto, Yoshimasa Tsuruoka, Richard Socher, et~al.
\newblock 2017.
\newblock A joint many-task model: Growing a neural network for multiple nlp
  tasks.
\newblock In {\em Proceedings of the 2017 Conference on Empirical Methods in
  Natural Language Processing}, pages 1923--1933.

\bibitem[\protect\citename{Huang \bgroup et al.\egroup
  }2015]{huang2015bidirectional}
Zhiheng Huang, Wei Xu, and Kai Yu.
\newblock 2015.
\newblock Bidirectional {LSTM-CRF} models for sequence tagging.
\newblock {\em arXiv preprint arXiv:1508.01991}.

\bibitem[\protect\citename{Ioffe and Szegedy}2015]{ioffe2015batch}
Sergey Ioffe and Christian Szegedy.
\newblock 2015.
\newblock Batch normalization: Accelerating deep network training by reducing
  internal covariate shift.
\newblock In {\em International conference on machine learning}, pages
  448--456.

\bibitem[\protect\citename{Kingma and Ba}2014]{kingma2014adam}
Diederik~P Kingma and Jimmy Ba.
\newblock 2014.
\newblock Adam: A method for stochastic optimization.
\newblock {\em arXiv preprint arXiv:1412.6980}.

\bibitem[\protect\citename{Lample \bgroup et al.\egroup
  }2016]{lample2016neural}
Guillaume Lample, Miguel Ballesteros, Sandeep Subramanian, Kazuya Kawakami, and
  Chris Dyer.
\newblock 2016.
\newblock Neural architectures for named entity recognition.
\newblock In {\em NAACL-HLT}, pages 260--270.

\bibitem[\protect\citename{LeCun \bgroup et al.\egroup
  }1989]{lecun1989backpropagation}
Yann LeCun, Bernhard Boser, John~S Denker, Donnie Henderson, Richard~E Howard,
  Wayne Hubbard, and Lawrence~D Jackel.
\newblock 1989.
\newblock Backpropagation applied to handwritten zip code recognition.
\newblock {\em Neural computation}, 1(4):541--551.

\bibitem[\protect\citename{Ling \bgroup et al.\egroup }2015]{ling2015finding}
Wang Ling, Chris Dyer, Alan~W Black, Isabel Trancoso, Ramon Fermandez, Silvio
  Amir, Luis Marujo, and Tiago Luis.
\newblock 2015.
\newblock Finding function in form: Compositional character models for open
  vocabulary word representation.
\newblock In {\em Proceedings of the 2015 Conference on Empirical Methods in
  Natural Language Processing}, pages 1520--1530.

\bibitem[\protect\citename{Liu \bgroup et al.\egroup }2018]{liu2017empower}
Liyuan Liu, Jingbo Shang, Frank Xu, Xiang Ren, Huan Gui, Jian Peng, and Jiawei
  Han.
\newblock 2018.
\newblock Empower sequence labeling with task-aware neural language model.
\newblock {\em AAAI}.

\bibitem[\protect\citename{Luo \bgroup et al.\egroup }2015]{luo2015joint}
Gang Luo, Xiaojiang Huang, Chin-Yew Lin, and Zaiqing Nie.
\newblock 2015.
\newblock Joint entity recognition and disambiguation.
\newblock In {\em EMNLP}, pages 879--888.

\bibitem[\protect\citename{Ma and Hovy}2016]{ma2016end}
Xuezhe Ma and Eduard Hovy.
\newblock 2016.
\newblock End-to-end sequence labeling via {B}i-directional {LSTM-CNNs-CRF}.
\newblock In {\em ACL}, volume~1, pages 1064--1074.

\bibitem[\protect\citename{Marcus \bgroup et al.\egroup
  }1993]{marcus1993building}
Mitchell~P Marcus, Mary~Ann Marcinkiewicz, and Beatrice Santorini.
\newblock 1993.
\newblock Building a large annotated corpus of english: The penn treebank.
\newblock {\em Computational linguistics}, 19(2):313--330.

\bibitem[\protect\citename{Mikolov \bgroup et al.\egroup
  }2013]{mikolov2013distributed}
Tomas Mikolov, Ilya Sutskever, Kai Chen, Greg~S Corrado, and Jeff Dean.
\newblock 2013.
\newblock Distributed representations of words and phrases and their
  compositionality.
\newblock In {\em Advances in neural information processing systems}, pages
  3111--3119.

\bibitem[\protect\citename{Passos \bgroup et al.\egroup
  }2014]{passos2014lexicon}
Alexandre Passos, Vineet Kumar, and Andrew McCallum.
\newblock 2014.
\newblock Lexicon infused phrase embeddings for named entity resolution.
\newblock In {\em CoNLL}, pages 78--86.

\bibitem[\protect\citename{Pennington \bgroup et al.\egroup
  }2014]{pennington2014glove}
Jeffrey Pennington, Richard Socher, and Christopher Manning.
\newblock 2014.
\newblock Glove: Global vectors for word representation.
\newblock In {\em Proceedings of the 2014 conference on empirical methods in
  natural language processing (EMNLP)}, pages 1532--1543.

\bibitem[\protect\citename{Peters \bgroup et al.\egroup }2017]{peters2017semi}
Matthew Peters, Waleed Ammar, Chandra Bhagavatula, and Russell Power.
\newblock 2017.
\newblock Semi-supervised sequence tagging with bidirectional language models.
\newblock In {\em ACL}, volume~1, pages 1756--1765.

\bibitem[\protect\citename{Ratinov and Roth}2009]{ratinov2009design}
Lev Ratinov and Dan Roth.
\newblock 2009.
\newblock Design challenges and misconceptions in named entity recognition.
\newblock In {\em CoNLL}, pages 147--155.

\bibitem[\protect\citename{Rei}2017]{rei2017semi}
Marek Rei.
\newblock 2017.
\newblock Semi-supervised multitask learning for sequence labeling.
\newblock In {\em Proceedings of the 55th Annual Meeting of the Association for
  Computational Linguistics (Volume 1: Long Papers)}, volume~1, pages
  2121--2130.

\bibitem[\protect\citename{Reimers and Gurevych}2017a]{reimers2017optimal}
Nils Reimers and Iryna Gurevych.
\newblock 2017a.
\newblock Optimal hyperparameters for deep lstm-networks for sequence labeling
  tasks.
\newblock {\em arXiv preprint arXiv:1707.06799}.

\bibitem[\protect\citename{Reimers and Gurevych}2017b]{reimers2017reporting}
Nils Reimers and Iryna Gurevych.
\newblock 2017b.
\newblock Reporting score distributions makes a difference: Performance study
  of lstm-networks for sequence tagging.
\newblock In {\em Proceedings of the 2017 Conference on Empirical Methods in
  Natural Language Processing}, pages 338--348.

\bibitem[\protect\citename{Santos and Zadrozny}2014]{santos2014learning}
Cicero~D Santos and Bianca Zadrozny.
\newblock 2014.
\newblock Learning character-level representations for part-of-speech tagging.
\newblock In {\em Proceedings of the 31st International Conference on Machine
  Learning (ICML-14)}, pages 1818--1826.

\bibitem[\protect\citename{Srivastava \bgroup et al.\egroup
  }2014]{srivastava2014dropout}
Nitish Srivastava, Geoffrey~E Hinton, Alex Krizhevsky, Ilya Sutskever, and
  Ruslan Salakhutdinov.
\newblock 2014.
\newblock Dropout: a simple way to prevent neural networks from overfitting.
\newblock {\em Journal of Machine Learning Research}, 15(1):1929--1958.

\bibitem[\protect\citename{Strubell \bgroup et al.\egroup
  }2017]{strubell2017fast}
Emma Strubell, Patrick Verga, David Belanger, and Andrew McCallum.
\newblock 2017.
\newblock Fast and accurate entity recognition with iterated dilated
  convolutions.
\newblock In {\em Proceedings of the 2017 Conference on Empirical Methods in
  Natural Language Processing}, pages 2670--2680.

\bibitem[\protect\citename{Tieleman and Hinton}2012]{tieleman2012lecture}
Tijmen Tieleman and Geoffrey Hinton.
\newblock 2012.
\newblock Lecture 6.5-rmsprop: Divide the gradient by a running average of its
  recent magnitude.
\newblock {\em COURSERA: Neural networks for machine learning}, 4(2):26--31.

\bibitem[\protect\citename{Tjong Kim~Sang and
  Buchholz}2000]{tjong2000introduction}
Erik~F Tjong Kim~Sang and Sabine Buchholz.
\newblock 2000.
\newblock Introduction to the conll-2000 shared task: Chunking.
\newblock In {\em Proceedings of the 2nd workshop on Learning language in logic
  and the 4th conference on Computational natural language learning-Volume 7},
  pages 127--132. Association for Computational Linguistics.

\bibitem[\protect\citename{Tjong Kim~Sang and
  De~Meulder}2003]{tjong2003introduction}
Erik~F Tjong Kim~Sang and Fien De~Meulder.
\newblock 2003.
\newblock Introduction to the conll-2003 shared task: Language-independent
  named entity recognition.
\newblock In {\em HLT-NAACL}, pages 142--147.

\bibitem[\protect\citename{Toutanova \bgroup et al.\egroup
  }2003]{toutanova2003feature}
Kristina Toutanova, Dan Klein, Christopher~D Manning, and Yoram Singer.
\newblock 2003.
\newblock Feature-rich part-of-speech tagging with a cyclic dependency network.
\newblock In {\em Proceedings of the 2003 Conference of the North American
  Chapter of the Association for Computational Linguistics on Human Language
  Technology-Volume 1}, pages 173--180. Association for Computational
  Linguistics.

\bibitem[\protect\citename{Yang \bgroup et al.\egroup }2016]{yang2016multi}
Zhilin Yang, Ruslan Salakhutdinov, and William Cohen.
\newblock 2016.
\newblock Multi-task cross-lingual sequence tagging from scratch.
\newblock {\em arXiv preprint arXiv:1603.06270}.

\bibitem[\protect\citename{Yang \bgroup et al.\egroup }2017a]{yang2017neural}
Jie Yang, Yue Zhang, and Fei Dong.
\newblock 2017a.
\newblock Neural reranking for named entity recognition.
\newblock In {\em Proceedings of the International Conference Recent Advances
  in Natural Language Processing, RANLP 2017}, pages 784--792.

\bibitem[\protect\citename{Yang \bgroup et al.\egroup }2017b]{yang2017transfer}
Zhilin Yang, Ruslan Salakhutdinov, and William~W Cohen.
\newblock 2017b.
\newblock Transfer learning for sequence tagging with hierarchical recurrent
  networks.
\newblock In {\em ICLR}.

\bibitem[\protect\citename{Zeiler}2012]{zeiler2012adadelta}
Matthew~D Zeiler.
\newblock 2012.
\newblock Adadelta: an adaptive learning rate method.
\newblock {\em arXiv preprint arXiv:1212.5701}.

\bibitem[\protect\citename{Zhai \bgroup et al.\egroup }2017]{zhai2017neural}
Feifei Zhai, Saloni Potdar, Bing Xiang, and Bowen Zhou.
\newblock 2017.
\newblock Neural models for sequence chunking.
\newblock In {\em AAAI}, pages 3365--3371.

\end{thebibliography}

\end{document}